\pdfoutput=1
\documentclass{article}
\usepackage{spconf,amsmath,graphicx}

\usepackage{enumitem}
\setlist{nosep, leftmargin=14pt}

\usepackage{mwe} % to get dummy images

\usepackage{amssymb}
\usepackage{caption}
\usepackage{subcaption}
\usepackage{multirow}
\usepackage{xcolor}
\usepackage{amsthm}
\usepackage{cleveref}
\usepackage{bigstrut}

\usepackage{spconf}  % Ensure this is the right package

\title{Learning Covariance-based Multi-scale Representation of NeuroImaging Measures for Alzheimer Classification}

\name{Seunghun Baek$^{\star}$, Injun Choi$^{\star}$, Mustafa Dere$^{\dagger}$, 
	Minjeong Kim$^{\ddagger}$, Guorong Wu$^{\dagger}$, Won Hwa Kim$^{\star}$}

\address{
	$^{\star}$ Pohang University of Science and Technology, Pohang, South Korea \\
	$^{\dagger}$ University of North Carolina at Chapel Hill, Chapel Hill, USA \\
	$^{\ddagger}$ University of North Carolina at Greensboro, Greensboro, USA
}

\begin{document}
	
	\ninept  % Required for spconf.sty compatibility
	
	\maketitle

\begin{abstract}

%Deep Neural Network (DNN) has successfully solved numerous tasks in image analysis. %successfully in a data-driven manner.
% When the input and target have a skewed relationship, %non-linear relationship?
% stacking several hidden layers can locate the data in high-dimensional latent space to solve tasks. %...?
%However, adopting several layers 
Stacking excessive layers in DNN %increases %the model size resulting 
results in highly underdetermined system when training samples are limited, which is very common in medical applications. 
% However, when only limited samples are available, demanding increase in model size by stacking layers makes itself as abominably underdetermined system.
In this regard, we present a framework capable of 
%revealing multi-dimensional relationship 
deriving an efficient high-dimensional space with reasonable increase in model size.
This is done by utilizing a transform (i.e., convolution) that leverages scale-space theory with covariance structure. 
The overall model trains on this transform together with a downstream classifier (i.e., Fully Connected layer) 
to capture the optimal multi-scale representation of the original data which corresponds to task-specific components in a dual space. 
%consists of %Together with a downstream classifier, 
%Along the covariance structure from the data, 
%band-pass kernels via trainable scales are stretched or shrunk to capture various task-specific components in a dual space.
Experiments on neuroimaging measures from Alzheimer's Disease Neuroimaging Initiative (ADNI) study 
show that our model performs better and converges faster than conventional models 
even when the model size is significantly reduced. 
The trained model is made interpretable using gradient information over the multi-scale transform 
to delineate personalized AD-specific regions in the brain. 

\end{abstract}
%
% \begin{keywords}
% One, two, three, four, five
% \end{keywords}
%

% What is the problem?
% Why is it interesting and important?
% Why is it hard? (E.g., why do naive approaches fail?)
% Why hasn't it been solved before? (Or, what's wrong with previous proposed solutions? How does mine differ?)
% What are the key components of my approach and results? Also include any specific limitations.
% \vspace{-5pt}
\section{Introduction}
\vspace{-10pt}
Literature in the past decades clearly demonstrates that 
Artificial Neural Network (ANN) has brought significant improvements in 
various prediction tasks for image data~\cite{Image_Seg_survey,Medical_image_survey}. 
However, it is also shown that they often require 
unprecedentedly large amount of data when the ANNs are deeply stacked 
to learn highly non-linear functions in a data-driven manner~\cite{DNN_survey,Medical_survey}. 
Such an issue limits the use of powerful ANN methods 
for medical image analyses where sample sizes are 
limited due to several difficulties in data acquisition~\cite{Medical_image_survey2}. 

%At the core of the issues from conventional ANNs, 
This critical issue is caused from the intrinsic architecture of ANN: the model becomes too flexible and complex 
even with only a few stacked hidden layers such that relying solely on the large-scale  data is inevitable. 
Therefore, the deeply stacked ANNs, i.e., Deep Learning (DL) approaches, are often used as a black box 
without much understanding of the behavior mechanism. 
Several tries \cite{expalinable_AI,expalinable_AI2} have effectively identified on which variables the model cares to come up with a specific decision, e.g., Class Activation Map (CAM)~\cite{cam} and Grad-CAM~\cite{grad_cam}, 
but such delineations come from post-hoc analysis of a trained model for a specific target task. 

\begin{figure}[t]
    \centering
    \centerline{\includegraphics[width=\columnwidth]{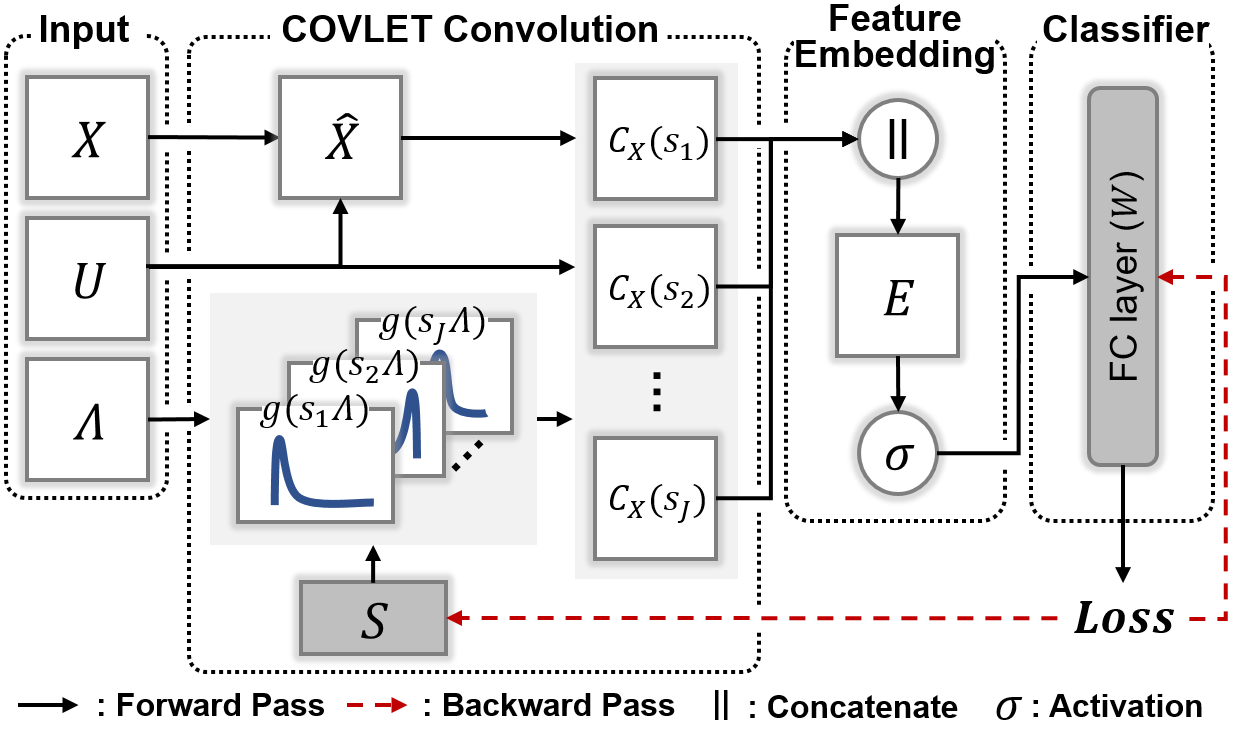}}
    \vspace{-5pt}
    \caption{\small Overall scheme of our multi-scale learning network. Input $X$ is transformed to a high-dimensional space with kernels $g(s)$ and Principal Components $U$ (i.e., convolution) and fed to a downstream classifier (solid line). 
    The $S$ and classifier are trained to obtain the optimal task-specific multi-scale representation (dashed line).
    % The size of boxes reflects the shape of each variable.
    } 
    \label{fig:model}
    \vspace{-15pt}
\end{figure}

%At the core of various deep ANN models, what they 
Essentially, what DL models learn is 
a {\em transformation} of the data into a latent space where the transformed representation is optimized for a target task~\cite{fan_paper}. 
Most of the DL approaches learn this transform in a non-parametric data-driven manner that results in estimating exhaustive parameters. 
In this regime, we propose to utilize parametric kernels (i.e., filter) as in wavelet transform~\cite{wavelet}, which is formulated 
as band-pass filtering in the Fourier space (i.e., frequency space). 
The bottleneck is that typical data (e.g., data matrix) are not defined in time and there is no notion of order of variables, and we tackle this issue using the covariance structure. % from the data. 
Leveraging the transform in \cite{covlet} which defines multi-scale representations using parametric kernels in the space spanned by eigenvectors of a covariance matrix, we design a novel neural network that trains on the ``scales'' of filters instead of learning the kernel itself. 
%Such a framework 
It learns 
the optimal transform by a simple convolutional filtering, %process as a convolution, 
%with significantly reduced number of parameters to learn, 
yielding a suitable multi-scale representation for a target task such as classification. 
%The trained transform as a convolution is analogous to the wavelet transform which is well studied in signal processing. 

To this end, our work brings the following contributions: 
1) we construct a neural network architecture that learns a multi-scale representation of tabular data via its covariance structure, 
2) our model is light and trains efficiently even with small number of samples, 
3) the proposed model is validated on various imaging measures %of different modalities 
from Alzheimer's Disease Neuroimaging Initiative (ADNI) study 
for classification of diagnostic labels for Alzheimer's Disease (AD). 
The experiments on region-wise cortical thickness from magnetic resonance image (MRI) and tau from positron emission tomography (PET) 
show that our model with a simple convolution layer can distinguish AD-related groups with high precision outperforming conventional ANN with similar architecture.

% 

% While there are active studies on training with limited samples in other data types\cite{}, tabular data has not been dealt with much.
% The main obstacle that makes tabular dataset out-of-interest is lack of commonly shared structures and biases.
% This is why augmentation can't be a simple solution to tabular data shortage problems.
% \cite{tabular_self_superv} came up with self-supervised learning, but 

% Unlike fancy baselines that works on other data types uses, tabular dataset still uses deterministic algorithm (i.e. SVM) as baselines.

%\vspace{-15pt}
\vspace{-10pt}
\section{METHODS}
\label{sec:format}
\vspace{-10pt}

As introduced in Fig.~\ref{fig:model}, we construct an efficient ANN utilizing a transform with parametric kernels summarized by scales and train on the scales only. 
%Our proposed model is introduced in Fig.~\ref{fig:model}, and the methodological details are described below. 

%Fig.\ref{fig:model} visualizes the overall process of our network.  
%\subsection{Tabular Data Notation}
\vspace{-7pt}
\subsection{Prelim: Covariance-based Multi-scale Transform}
\vspace{-5pt}
\label{ssec:tabular_data}
% %$\chi = (\mathcal{F},X,y)$ denotes a type of specific data structure, where $\mathcal{F}$ specifies the raw feature type space, X is the feasible instance space, and y is the target space.
% Given $X_{}$ denotes a type of specific data structure, 
% where $\mathcal{F}$ specifies the raw feature type space, X is the feasible instance space, and y is the target space.
% In a tabular dataset of $\chi$ type, 
% an instance $x\in\mathbb{R}^p$ in $X$ is defined as an $p$-element vector representing $p$ scalar raw features in $\mathcal{F}(n=|\mathcal{F}|)$.
% As \cite{DANETs}, we assume that there are some underlying feature groups in a
% tabular data structure, and the features in a group are correlative and target-relevant.

%\subsection{Frequency Analysis of Tabular Data}
%\label{ssec:frequency}
Let $X\in\mathbb{R}^{p\times n}$ be a standardized (zero-mean) feature matrix with $n$ samples with $p$-dimensional feature. 
Its covariance matrix $\Sigma_{p\times p}=\frac{1}{n}XX^T$ shows how each variable is related to each other. %across different variables. 
%in multivariate normal distribution setting.
Because a covariance matrix is real and symmetric, 
it has a complete set of orthonormal eigenvectors $U = [u_1|u_2|...|u_p]$ and corresponding positive definite eigenvalues %$\Lambda = diag(\lambda_0,...,\lambda_p)$ where, 
$0 < \lambda_1 \leq ... \leq \lambda_p$.
%Likewise in PCA, 
These eigenvectors, known as Principal Components (PC), define an orthonormal subspace where 
$X$ can be projected onto with minimal loss of information \cite{pca}.  %into a new subspace.
%Using all eigenvectors together, 
Using $U$, $X$ can be transformed as a signal $\hat X = U^TX$ in a dual space (i.e., an analogue of the frequency space), where an eigenvector that corresponds to a larger eigenvalue captures components of $X$ with high variation. 
As in conventional frequency analysis, one can apply a filter $g(\cdot)$ on $\hat X$ as 
%$g(\Lambda)U^TX$ where $\Lambda = diag(\lambda_0,...,\lambda_p)$.
$g(\Lambda)\hat X$ where $\Lambda = diag(\lambda_1,...,\lambda_p)$. 
%and transform it back to the original space 
%to obtain a new representation. 
When the $g(\cdot)$ is a band-pass filter of a specific scale $s$, 
the formulation becomes a Wavelet-like transform with coefficients 
%whose coefficients can be obtained as 
\begin{equation}\small
\vspace{-2pt}
    C_X(s) = Ug(s\Lambda)U^TX
\label{eq:C_X(S)}
\vspace{-2pt}
\end{equation}
that yields a multi-resolution representation of $X$ on the covariance structure of $\Sigma$. 
This transform was introduced as Covariance-based Wavelet-like (COVLET) transform in \cite{covlet}.

%In this dual space, one can analyze $\hat x$ as an analogue of the data in the frequency space as in conventional 
%In the dual space, the signal $\hat x$ enables to handle given tabular data $X$ likewise in image, audio and graph.

% In the dual space of $X$, we adopt a cubic spline kernel $g(\cdot)$ from \cite{SGWT} which was basically designed for graph frequency space.
% A seamless and differentiable kernel is formulated as
% \begin{equation}
%     g(x) =
%     \left\{
%     \begin{array}{cl}
%         x,& 0<x<1 \\
%         -5+11x-6x^2+x^3,& 1\leq x\leq 2 \\
%         4x^{-2},& 2<x
%     \end{array}
%     \right ..
% \label{eq:kernel}
% \end{equation}

% \begin{figure}[h]
%     \centering
%     \centerline{\includegraphics[width=\columnwidth]{figure/scale_shape.png}}
%     \caption{\small Shape of kernel $g(sx)$ with various scale parameter $s$}
%     \label{fig:scale_shape}
% \end{figure}
% By scaling a input $x$ of the kernel with scale parameter $s$, the kernel shape is changed in terms of $x$ and decides where to mainly highlights as shown in \figurename{\ref{fig:scale_shape}}.

%This squeezed/stretched kernels can attach a different weight to each frequency component which is just $U^TX$ by being multiplied as $diag(g(s\Lambda))U^TX$. Returning transformed signal back to tabular data domain, $Udiag(g(s\Lambda))U^TX$ can be interpreted as an embedding of $X$.

\vspace{-7pt}
\subsection{Learning Adaptive Kernel with Scales}
\vspace{-5pt}
\label{sec:kernel}
%Because this kernel is differentiable with respect to $s$, adaptive scales for a given task can be trained end-to-end from the task loss.

Given a set of positive and trainable scales $S=[s_1, s_2,...,s_J]$
where $J=|S|$, each $s_i$ yields an embedding of $X$ as $C_X(s_i) = Ug(s_i\Lambda)U^TX$. 
%so-called CMD$_i$ (Covlet Multi-scale Descriptor). 
With $S\in\mathbb{R}^{J}$ and $X\in \mathbb{R}^{p\times n}$, a set of embeddings is obtained and concatenated together as $E = [C_X(s_1),C_X({s_2}),...,C_X({s_J})] \in \mathbb{R}^{(J\times p)\times n}$ which is a mapping of the $X$ onto a higher-dimensional space. 
To capture target task-relevant frequency components, 
the $S$ is trained to minimize a task-specific loss function, e.g., cross-entropy for classification. 
%our model adopt

% \begin{lemma}
% If the filter $g(\cdot)$ in \eqref{eq:C_X(S)} is differentiable with respect to $s_i$, 
% % then a gradient of $s_i$ can be trained with a task-specific loss function $L_{task}$.
% then the representation \eqref{eq:C_X(S)} can be optimized to minimize $L_{task}$a gradient of $s_i$ can be trained with a task-specific loss function $L_{task}$.
% \label{lemma1}
% \end{lemma}
% \begin{proof}
% As widely known, a task-specific loss can be back-propagated toward the neural network's input, i.e. $C_X(S_i)$ in our model.
% Assuming $g(\cdot)$ can be differentiated toward its input as $g'(\cdot)$,
% $s_i$ can be trained in a data-driven way as we formulate a gradient of $L_{task}$ toward $s_i$ as:
% \begin{equation}
%     \frac{\partial L_{task}}{\partial s_i} = \frac{\partial L_{task}}{\partial C_X(s_i)}\frac{\partial C_X(s_i)}{\partial g(s_i\Lambda)}\frac{\partial g(s_i\Lambda)}{\partial s_i\Lambda} \frac{\partial s_i\Lambda}{\partial s_i}.
% \label{eq:grad_s}
% \end{equation}
% \end{proof}

If the filter $g(\cdot)$ in \eqref{eq:C_X(S)} is differentiable with respect to $s_i$, 
% then a gradient of $s_i$ can be trained with a task-specific loss function $L_{task}$.
then the representation from \eqref{eq:C_X(S)} can be optimized according to a task-specific loss $L_{task}$. 
This is because the derivative $\frac{\partial L_{task}}{\partial s_i}$ can be achieved via chain rule as
%a gradient of $s_i$ can be trained with a task-specific loss function $L_{task}$.
%As widely known, a task-specific loss can be back-propagated toward the neural network's input, i.e. $C_X(S_i)$ in our model.
%Assuming $g(\cdot)$ can be differentiated toward its input as $g'(\cdot)$,
%$s_i$ can be trained in a data-driven way as we formulate a gradient of $L_{task}$ toward $s_i$ as:
{\small
\vspace{-3pt}
\begin{align}
    \frac{\partial L_{task}}{\partial s_i} &= \frac{\partial L_{task}}{\partial C_X(s_i)}\frac{\partial C_X(s_i)}{\partial g(s_i\Lambda)}\frac{\partial g(s_i\Lambda)}{\partial s_i\Lambda} \frac{\partial s_i\Lambda}{\partial s_i}\\
     &= \frac{\partial L_{task}}{\partial C_X(s_i)}\frac{\partial C_X(s_i)}{\partial g(s_i\Lambda)}g'(s_i\Lambda) \Lambda. 
\label{eq:grad_s}
\vspace{-2pt}
\end{align}}%
%Using the \eqref{eq:grad_s} above, 
Eq. \eqref{eq:grad_s} says that the loss can be back-propagated via a gradient descent method throughout the neural network 
to update $C_X(s_i)$ if $g'(\cdot)$ exists. 
%i.e. $C_X(s_i)$ in our model, . 
Such optimization achieves the optimal $s_i$ that corresponds to a specific scale of the band-pass filter. 
Learning multi-variate $S$ yields the optimal multi-scale representation $C_X(s_i), i\in\{1,\cdots,J\}$ of $X$. 
%of the original $X$ along the covariance structure. 
%Assuming $g(\cdot)$ can be differentiated toward its input as $g'(\cdot)$, $s_i$ can be trained in a data-driven way as we formulate a gradient of $L_{task}$ toward $s_i$ as:
%In our method, as claimed in \cref{lemma1}, we adopt differentiable band-pass filter $g(\cdot)$ to capture task-relevant frequency components with.

% Our model adopt differentiable spline cubic function from \cite{SGWT} as $g(\cdot)$, where $\partial g(\cdot) = g'(\cdot)$.
To make use of all $C_X(s_i)$ in downstream task, we concatenated these multi-scale representations into $E$.
%This representation $E$ can be treated as an output of a single hidden layer whose weight matrix is $W\in\textit{}{R}^{(J\times p)\times p}$.
Because $E$ is a %can be represented as 
linear transform of $X$, 
an activation function $\sigma$ is required to make our network non-linear toward $X$ as $\sigma(E)$.
We feed this non-linear multi-scale representation $\sigma(E)$ to classifier (i.e., fully connected layer with weights $W$) at the end, 
which produces a class-wise prediction $\hat y$. 
The overall architecture trains on $W$ and $S$ 
to minimize multi-class cross-entropy between the $\hat y$ and the ground truth $y$. 
% \begin{equation}
%     L(s,W) = 
% \end{equation}

% \textcolor{red}{\textbf{Interpretation.} % Preliminary?
% The trained model is made interpretable with Grad-CAM \cite{grad_cam}.
% % 'differentiable' model?
% When a sample is fed to trained model, with back-propagation from the model's prediction (i.e. class label), 
% gradients can be computed toward every layer's input.
% Regarding each gradient as an importance of corresponding input in the layer, 
% weighted combination with layer's input followed by ReLU can tell us which input is positively correlated with model's prediction.}

\textbf{Model Interpretation.} % Preliminary?
%We revise the Grad-CAM \cite{grad_cam} to summarize high-dimensional feature due to multi-scale transform so that we can identify which of the variable was highly responsible to predict a specific label. 
% 'differentiable' model?
When a sample is fed to a trained model, with a back-propagation of the model's prediction (i.e. class label), 
gradients are computed toward every layer's input.
Considering each gradient as an importance measure of corresponding input in the layer, 
weighted combination with layer's input followed by Rectified Linear Unit (ReLU) tells us which variable is related to the model's prediction, and this is known as Grad-CAM \cite{grad_cam}.
% This correlation is known as CAM (class activation map).
In our framework, Grad-CAM can be obtained for every multi-scaled representation $C_X(s_i)$.
To figure out the ROI-wise influence across the scales, the Grad-CAM is averaged across the scale 
to yield a measure $M$ for $k$-th input variable on $c$-th class as
%To figure out the variable-wise influence, we averaged these results over scales.
%This enables us to assign a single Grad-CAM value $M$ for $k$-th input variable on $c$-th class as
\begin{equation}\small
\vspace{-3pt}
    M^c(k) = \sum_{i=1}^J ReLU(\frac{\partial \hat y_c}{\partial C_{X_k}(s_i)}\cdot C_{X_k}(s_i))
\label{eq:gradcam}
\vspace{-1pt}
\end{equation}
where 
$\hat y_c$ is a class-wise prediction and $C_{X_k}(s_i)$ denotes embedding of $k$-th input variable with $s_i$. 
%Using $M$, we can 
The $M$ identifies ROI-wise effect of a specific label for each subject.

% \subsection{Parameter Comparisons}
% In this section, we compare the number of parameters in a conventional Multi-layer perceptron (MLP) with our framework. 
% Considering that the current architecture consists of 2 layers, i.e., one with COVLET transform and 
% the other acting as a classifier, 
% we compare it with a typical 2-layer MLP. 
% %We feed this embedding into a downstream single layered classifier to compare a number of parameters with 
% %Multi-Layer Perceptron with two layers (2-MLP). 
% The Two models are formulated as
% \begin{align}
%     %\begin{array}{rcl}
%         \text{Ours} &=W_C\sigma([C_X(s_1),C_X({s_2}),...,C_X({s_J})]) \\
%         \text{2-MLP}&=W_{mlp}\sigma(W_{mlp}X)
%     %\end{array},
%     \label{eq:model_formula}
% \end{align}
% where $\sigma$ is activation function, $W_1, W_2, W_3$ are weights of linear layers. 
% When output's dimension is $k$, our model has trainable parameter $S\in\mathbb{R}^{c}$, and $W_1\in\mathbb{R}^{k\times(c\times p)}$. 
% In 2-MLP, $W_1\in\mathbb{R}^{k\times p}$ and $W_2\in\mathbb{R}^{p\times p}$ consist the model parameters. Total number of parameters in each model is
% \begin{equation}
%     \begin{array}{rcl}
%         Params_{ours} &=& k\times c\times p + c\\
%         Params_{2-MLP} &=& k\times p + p\times p
%     \end{array}.
% \end{equation}
% Our model has advantage if $c<\frac{p^2+kp}{1+kp}$.
% In most classification tasks, input dimension $p$ is much greater than $k$, which approximates this condition to $c < p$.
\vspace{-10pt}
\section{EXPERIMENTAL RESULTS}
\label{sec:experiment}
\vspace{-10pt}
In this section, we perform two experiments on ADNI data 
to validate the performance and efficiency of our method. 

\begin{table}[!b]
    \centering
    \vspace{-15pt}
    \small
    \caption{Demographics of ADNI dataset}
    \vspace{-10pt}
    \renewcommand{\arraystretch}{0.73}
    \scalebox{0.70}{
    \begin{tabular}{|c|c||c|c|c|c|}
        \hline
        Biomarker & Category & CN & EMCI & LMCI & AD \bigstrut\\
        \hline
        \multirow{3}{*}{\shortstack{Cortical\\Thickness}} & \# of Subjects & 844 & 490 & 250 & 240 \bigstrut\\ \cline{2-6}
        & Age (mean, std) & 74.1$\pm$8.1 & 71.3$\pm$7.5 & 72.0$\pm$7.7 & 74.1$\pm$8.1 \bigstrut\\ \cline{2-6}
        & Gender (M/F) & 490/354 & 282/208 & 148/102 & 149/91 \bigstrut\\ \cline{2-6}
        \hline
        \multirow{3}{*}{\shortstack{Tau\\protein}} & \# of Subjects & 237 & 186 & 105 & 85 \bigstrut\\ \cline{2-6}
        & Age (mean, std) & 72.0$\pm$5.9 & 70.1$\pm$7.0 & 70.6$\pm$7.7 & 72.6$\pm$8.0 \bigstrut\\ \cline{2-6}
        & Gender (M/F) & 119/118 & 110/76 & 61/44 & 43/42 \bigstrut\\ \cline{2-6}
        \hline
    \end{tabular}}
    \label{tab:ADNI}
\end{table}

\begin{table*}[!t]
\caption{Performance comparison of baselines and our model (with 16 scales) on each two classification tasks from two different biomarkers. 
The number of parameters in each model is reported together (except Linear SVM).}
\vspace{-5pt}
\label{tab:classification_performance}
\centering
\scalebox{0.62}{
\setlength{\tabcolsep}{1.4em}
\renewcommand{\arraystretch}{0.8}
\begin{tabular}{|c|c||c|c|c|c||c|c|c|c|}
\hline
\multirow{2}{*}{Biomarker} & \multirow{2}{*}{Methods} & \multicolumn{4}{c||}{CN vs. EMCI} & \multicolumn{4}{c|}{CN vs. EMCI vs. LMCI vs. AD}\bigstrut\\
\cline{3-10}
 &  & \# Params & Accuracy & Recall & Precision & \# Params & Accuracy & Recall & Precision\bigstrut\\
\hline\hline
\multirow{6}{*}{\shortstack{Cortical\\Thickness}} & Linear SVM & - & 0.738$\pm$0.026 & 0.734$\pm$0.031 & 0.725$\pm$0.027 & - & 0.742$\pm$0.017 & 0.726$\pm$0.022 & 0.739$\pm$0.027\bigstrut\\
\cline{2-10}
& 1-MLP & 0.3K & 0.738$\pm$0.023 & 0.728$\pm$0.034 & 0.723$\pm$0.027 & 0.6K & 0.683$\pm$0.014 & 0.655$\pm$0.018 & 0.669$\pm$0.013\bigstrut\\
\cline{2-10}
& 2-MLP$_{R}$ & 5.2K & 0.759$\pm$0.019 & 0.754$\pm$0.029 & 0.745$\pm$0.021 & 10.5K & 0.692$\pm$0.011 & 0.672$\pm$0.012 & 0.679$\pm$0.017\bigstrut\\
\cline{2-10}
& 2-MLP$_{I}$ & 25.9K & 0.765$\pm$0.045 & 0.759$\pm$0.053 & 0.750$\pm$0.048 &  26.2K & 0.702$\pm$0.013 & 0.681$\pm$0.010 & 0.691$\pm$0.016\bigstrut\\
\cline{2-10}
& Ours $(J$$=$$16)$ & 5.1K & \textbf{0.858$\pm$0.029} & \textbf{0.849$\pm$0.036} & \textbf{0.846$\pm$0.030} & 10.2K & \textbf{0.800$\pm$0.021} & \textbf{0.788$\pm$0.035} & \textbf{0.785$\pm$0.030}\bigstrut\\
\hline\hline
\multirow{6}{*}{\shortstack{Tau\\Protein}} & Linear SVM & - & 0.615$\pm$0.015 & 0.593$\pm$0.017 & 0.613$\pm$0.023 & - & 0.515$\pm$0.029 & 0.458$\pm$0.033 & 0.575$\pm$0.052\bigstrut\\
\cline{2-10}
& 1-MLP & 0.3K & 0.608$\pm$0.032 & 0.589$\pm$0.034 & 0.600$\pm$0.036 & 0.6K & 0.504$\pm$0.020 & 0.479$\pm$0.027 & 0.519$\pm$0.035\bigstrut\\
\cline{2-10}
& 2-MLP$_{R}$ & 5.2K & 0.648$\pm$0.030 & 0.635$\pm$0.033 & 0.642$\pm$0.034 & 10.5K & 0.540$\pm$0.019 & 0.514$\pm$0.020 & 0.563$\pm$0.047\bigstrut\\
\cline{2-10}
& 2-MLP$_{I}$ & 25.9K & 0.653$\pm$0.032 & 0.642$\pm$0.036 & 0.648$\pm$0.035 &  26.2K & 0.556$\pm$0.012 & 0.521$\pm$0.016 & \textbf{0.590$\pm$0.018}\bigstrut\\
\cline{2-10}
& Ours $(J$$=$$16)$ & 5.1K & \textbf{0.726$\pm$0.060} & \textbf{0.721$\pm$0.064} & \textbf{0.723$\pm$0.063} & 10.2K & \textbf{0.577$\pm$0.016} & \textbf{0.553$\pm$0.016} & 0.585$\pm$0.046\bigstrut\\
\hline
\end{tabular}}
\vspace{-10pt}
\end{table*}

\vspace{-7pt}
\subsection{ADNI Dataset}
\vspace{-5pt}
\label{ssec:dataset}
%We used magnetic resonance image (MRI), diffusion tensor image (DTI) and positron emission tomography (PET) imaging measures from the public Alzheimer's Disease Neuroimaging Initiative (ADNI) dataset for the experiment. %which is public dataset for longitudinal Alzheimer's disease study.
%Individual DTI were processed %by tractography pipeline to extract brain networks 
For the experiments, 
%we used magnetic resonance image (MRI) and positron emission tomography (PET) imaging measures from the public
we used MRI and PET imaging measures from the public 
%Alzheimer's Disease Neuroimaging Initiative (ADNI) dataset for the experiment.
ADNI study. 
The images were registered to Destrieux atlas~\cite{Parcellation} to obtain region-wise measures 
from 148 cortical regions and 12 sub-cortical regions, i.e., total of 160 regions. % based on Destrieux atlas~\cite{Parcellation}.
{\em Grey matter biomarkers} (i.e., cortical thickness from 148 cortical regions and grey matter volume from 12 sub-cortical regions) were derived from the MRI, and {\em tau} measure was obtained from the PET scans.  
%\textcolor{red}{XXX measures were obtained from the PET scans...}.
%With T-1 weighted magnetic resonance images (MRI), cortical thickness of each surface regions are calculated for each brain networks.
The dataset consists of 4 AD-specific progressive groups: Cognitively Normal (CN), Early Mild Cognitive Impairment (EMCI), Late Mild Cognitive Impairment (LMCI) and Alzheimer's Disease (AD). 
%% CANDIDATE 1
% These groups are reorganized to constitute several classification tasks to validate the proposed framework and identify AD-specific ROIs. 
The demographics of ADNI dataset is summarized in \tablename{ \ref{tab:ADNI}}.
%we used every groups except SMC and clustered these into two groups (AD, LMCI vs. EMCI, CN), three groups (AD vs. LMCI, EMCI vs. CN), and four groups(AD vs. LMCI vs. EMCI. vs CN).

\vspace{-7pt}
\subsection{Experiment Setup}
\vspace{-5pt}
%\subsection{Baseline Methods and Performance measures}
\label{ssec:experiment_setup}

{\bf Group Comparisons. } In our experiment, 
%the AD-relevant groups are %reorganized to 
we designed tasks of 2-way (CN vs. EMCI) %, 
%3-way (CN vs. EMCI+LMCI vs. AD) 
and 4-way (CN vs. EMCI vs. LMCI vs. AD) classifications to validate qualitative and quantitative performances of the proposed method.

\noindent {\bf Baselines.} 
% We adopted four baseline methods: 1) Linear Support Vector Machine (Linear SVM), 2) Multi-Layer Perceptron with one layer (1-MLP) and 3) with two layers (2-MLP). 
% For the 2-MLP, we tried two cases where the number of hidden nodes were consistent as the input dimension (2-MLP$_{I})$ or reduced to have similar number of parameters with our methods when $J=16$ (2-MLP$_{R}$).
We adopted four baseline methods: 1) Linear Support Vector Machine (Linear SVM), 2) Multi-Layer Perceptron with one layer (1-MLP) and 3,4) with two layers (2-MLP$_{I}$, 2-MLP$_{R}$).
For the 2-MLP, we tried two cases where the number of hidden nodes were consistent as the input dimension (2-MLP$_{I}$) or reduced to have similar number of parameters with our methods when $J$$=$$16$ (2-MLP$_{R}$).

\noindent{\bf Evaluation.} For unbiased and fair comparison, we carefully tuned baseline methods and used 5-fold cross validation (CV) in every experiment. 
For evaluation, we used mean accuracy, precision and recall across the CV.
Precision and recall were %calculated and 
averaged with equal importance for each class.

\noindent{\bf Training.} Due to the imbalance in class labels, we oversampled training dataset using ADASYN~\cite{ADASYN}. % before feeding the training data into the  prediction models. % including baseline methods and ours.  
%\subsection{Hyper-parameters}
For our method, %to capture every task-wise informative frequency components through trainable kernels, 
we tried various number of kernels $J\in[2, 64]$, 
with each scale initialized uniformly between [0.1,10] in $\log_{10}$-space. 
%An ablation study on $J$ is given
Performance analysis on the $J$ is given in Section \ref{ssec:model_behavior}.
%and $J$$=$$16$ yielded the best results. 
%positive scale parameters as a single vector $S\in\mathbb{R}^{16}$. 
Network weights were randomly initialized with He initialization~\cite{He_Initialization} 
%Cross-Entropy was used as a loss function. %as objective and 
and trained with AdamW~\cite{AdamW} optimizer %for back-propagation 
including both classifier $W$ and scale parameters $S$.
Learning rates for $W$ and $S$ were set separately within $[0.001, 0.03]$, but their effects were marginal. 
To prevent overfitting, weight decay at 0.01 was adopted for every linear layer.
Spline kernel in \cite{sgwt} was used for $g(\cdot)$ as it %as it is defined as 
behaves as a smooth band-pass filter.

\begin{figure*}[!t]
\centering
\includegraphics[width=.96\linewidth]{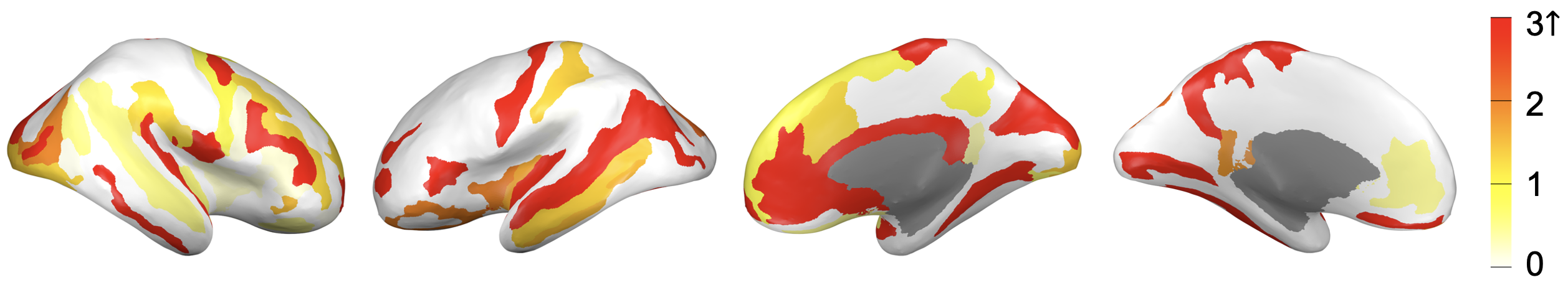}
\vspace{-10pt}
\caption{\small Visualization of $M$ on a random AD-sample that identifies personalized AD-specific ROIs. 
%that helps our model to decide the sample is in AD group using Grad-CAM~\cite{grad_cam}. 
Various AD-specific ROIs are identified as in \cite{AD_journal,AD_journal2,AD_journal3} including hippocampus, thalamus, amygdala and several temporal regions. Drawings were generated using BrainPainter~\cite{brain_painter}.}
\vspace{-15pt}
\label{fig:AD_CAM}
\end{figure*}

\vspace{-7pt}
\subsection{Performance Evaluation}
\vspace{-5pt}
\label{ssec: performance}
%In this section, we analyzed AD's progression of N=1824 participants in ADNI dataset as a classification task.
%Each participant provided cortical thickness and tau protein on every ROI as a column in tabular data respectively.
Our model is evaluated on two classification tasks 
as in Section \ref{ssec:experiment_setup} to demonstrate that it can classify different AD-related classes. 
%we conduct our experiments on three differently clustered groups on both biomarker to show our model's validity.
For every experiment, we reported performances in mean and standard deviation from 5-fold CV 
and they are summarized in \tablename{ \ref{tab:classification_performance}}.
%A tendency in performances exists over every experiment.
In every experiment, 
stacking one more layer on 1-MLP, i.e., 2-MLP$_R$ and 2-MLP$_I$, increased the performance where 
that of 2-MLP$_I$ was better than 2-MLP$_R$ possibly due to larger model size. 
%When stacking another linear layer on 1-MLP, which is simply 2-MLP$_R$ and 2-MLP$_I$, the performance increased as expected.
%In every experiment, 2-layered MLP with larger hidden nodes (2-MLP$_I$) yielded better results than one with less hidden nodes (2-MLP$_R$).
%Compared to 2-MLP$_I$ and 2-MLP$_R$, 
On top of the MLPs, our model with $J$$=$$16$ outperformed them with less or similar model sizes. 

\vspace{-7pt}
\subsubsection{2-way Classification: CN vs. EMCI}
\vspace{-5pt}
\label{sssec:2_way} 
We first demonstrate the performance of our model together with baselines on a binary classification task where CN and EMCI groups are distinguished. 
This is not an easy problem as the variation caused by AD in the preclinical stage is subtle. 

%Though the task may seem simple, it is not an easy problem as the variation caused by AD in the preclinical stage is subtle. 

\noindent \textbf{Cortical Thickness.} Linear SVM and single layer classifier (1-MLP) achieved almost 74\% in accuracy.  
Stacking one more linear layer (2-MLP$_R$, 2-MLP$_I$), 
% performance increased only $\sim$2\% in accuracy even though the number of parameters drastically increased.  %even with relatively excessive increase of parameters.
accuracy increased only $\sim$2\% even though model size increased drastically.  %even with relatively excessive increase of parameters.
However, when the convolution layer of our method with $J$$=$$16$ was added to the 1-MLP, 
it gained $\sim$12\% in accuracy even with significantly less number parameters than 2-MLP$_I$. 
%It achieved the highest accuracy, recall, and precision while keeping parameters less than multi-layer perceptron with two layers.
% Capacity..?
% When compared to 2-layered classifier with comparable model size (2-MLP$_R$), 
Compared to 2-MLP$_R$ with comparable model size,
our model with $J$$=$$16$ outperformed by $\sim$10\% in accuracy.

\noindent \textbf{Tau Protein.} 
%Due to the difference in sample size, overall performances using tau protein is lower than using cortical thickness. However, results was similar as in Cortical Thickness. 
We observed similar accuracy patterns in Tau as in the cortical thickness analysis. 
While 2-layered classifiers (2-MLP$_R$, 2-MLP$_I$) achieved nearly 4\% increase in accuracy compared to single layer classifier, our model with $J$$=$$16$ even outperformed 2-layered classifiers by 7\%.

% \subsubsection{AD vs. MCI vs. CN}
% \label{sssec:3_way}
% In this experiment, EMCI, LMCI are merged into a MCI group to design 
% a 3-way classification. 
% %%%%%%%%%
% % Is this harder than CN vs EMCI..?
% % What makes this task special..?
% %%%%%%%%%
% We validated the performance of our model as we did in CN vs. EMCI.

% \textbf{Cortical Thickness.} 
% Single layer classifier (1-MLP) performed about 70\% in accuracy. 
% In 2-MLP$_R$ and 2-MLP$_I$, each performance only increased only $\sim1\%$ and $\sim2\%$ in accuracy.
% In this experiment, Linear SVM performed better than 2-MLP$_R$ and 2-MLP$_I$.
% Still our model with $J=16$ achieved the highest performances in accuracy, recall, and precision.

% \textbf{Tau protein.}
% Our proposed model with $J=16$ outperformed other baselines (Lienar SVM, 1-MLP, 2-MLP$_R$, and 2-MLP$_I$) by nearly 8\%, 12\%, 6\%, 4\% respectively.

\vspace{-7pt}
\subsubsection{4-way Classification: CN vs. EMCI vs. LMCI vs. AD}
\vspace{-5pt}
\label{sssec:4_way}
%We set each class as a group to validate our model's performance on much harder task.
% We extend the experiment to 4-way classification to validate our model's performance with a more difficult case.
We extend the experiment to validate our model with a more difficult case.
% performance increased only $\sim$2\% in accuracy even though the number of parameters drastically increased.  %even with relatively excessive increase of parameters.
%As our model performed better in relatively easier task with less parameters, one can be suspicious toward our model's validity on much complicated tasks.
%In 4-way classification, basically it is harder than 2-way and 3-way classifications in that 
Here, the expected accuracy with a random guess is only 25\% and guessing with the majority class (i.e., CN) is $46.8\%$ for cortical thickness and $38.7\%$ for tau.
%Moreover, difficulty in distinguishing detailed every AD-relevant groups makes this 4-way classification more confusing.
%Here we validate our model is not just compressing the linear layer, but also better at mapping $X$ to task-relevant embedding space. 

\noindent \textbf{Cortical Thickness.} 
Classifier with MLPs (1-MLP, 2-MLP$_R$, 2-MLP$_I$) %did not perform well even with two layers.
achieved $\sim$70\% in accuracy even worse than Linear SVM which reached 74\% in accuracy.
Even in this complicated task, our model with $J$$=$$16$ yielded $80\%$ in accuracy surpassing other 2-MLP baselines over 10\% with less number of parameters. Precision and recall were around $0.78$ demonstrating that the model worked reasonably well. 

\noindent \textbf{Tau Protein.}
Linear SVM and 1-MLP achieved only about 50\% in accuracy.
While 2-MLP$_R$ and 2-MLP$_I$ outperformed 1-MLP in the accuracy by 4 to 5\%, 
Ours ($J$$=$$16$) surpassed them. 
Overall performances were not as good as in cortical thickness. 
% This may be because tau measures show change in the early stages of AD~\cite{tau_early_stage}, therefore it may not 
This may be because tau measures vary in the early stages of AD~\cite{tau_early_stage}, which may not 
be a suitable biomarker to characterize later stages of MCI and AD.

\vspace{-7pt}
\subsection{Interpretation from Trained Model}
\vspace{-5pt}
%As we tested our model on ADNI data, interpreting the model's behavior can help us to understand which ROIs are highly correlated to AD-progress.
%With the clinical backgrounds~\cite{AD_journal}, we validated our model as a cross-check.
In Fig. \ref{fig:AD_CAM}, we visualize Grad-CAM result using Eq.~\eqref{eq:gradcam} from Section \ref{sec:kernel}. 
This result was derived by inputting cortical thickness from a randomly selected AD sample into our pre-trained model with 4-way classification setting. 
It delineates which ROIs the model considered importantly to classify it as an AD sample, 
which include
%We fed cortical thickness data from one of AD-labeled samples into our best performing models on 4-way classification setting.
%With Grad-CAM~\cite{grad_cam}, we figured out which ROIs model took in consideration,  and visualized in \figurename{ \ref{fig:AD_CAM}}.
%Those ROIs mainly includes 
\textit{superior temporal, hippocampus, thalamus, amygdala} and many others which are %already verified in \cite{AD_journal}.
very well-known to be AD-specific in other literature \cite{AD_journal, AD_journal2, AD_journal3}.

\vspace{-7pt}
\subsection{Discussions on Model Behavior}
\begin{figure}[!h]
    \vspace{-5pt}
    \centering
    \includegraphics[width=0.95\linewidth, height = 3.8cm]{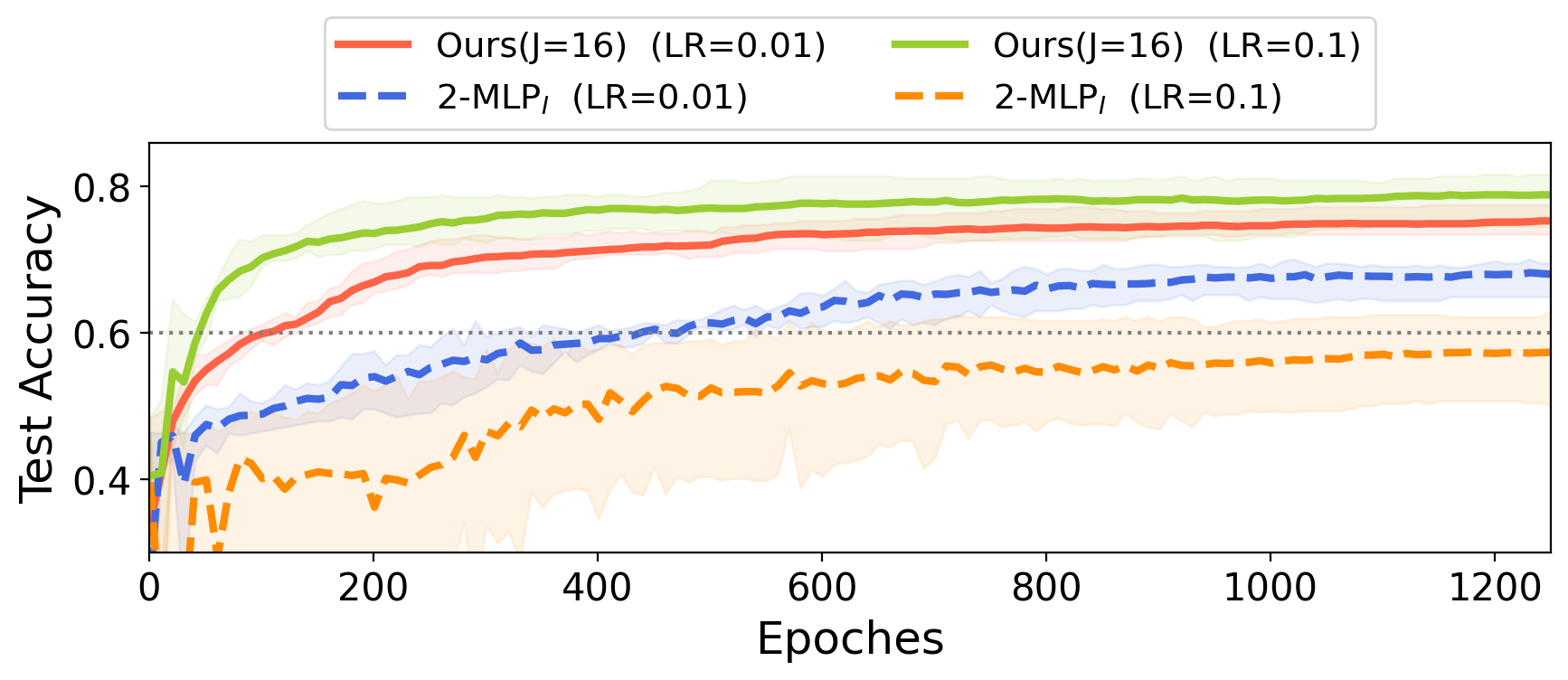} 
    \vspace{-5pt}
    \caption{\small Comparisons of mean test accuracy from our model and 2-MLP$_I$ on 4-way classification with cortical thickness. Our model reaches 0.6 significantly faster than 2-MLP$_I$. Measures are computed from 5-fold CV (shaded areas are range of the test accuracy). 
    %The test accuracy on 4-way classification using cortical thickness under same learning rate in our model and 2-MLP$_I$. The curve is the mean and shaded areas are range of test accuracy over 5-fold runs.
    }
    \label{fig:convergence_same_lr}
    %\vspace{-15pt}
\end{figure}

\vspace{-5pt}
\label{ssec:model_behavior}
\textbf{Convergence.} In \tablename{ \ref{tab:classification_performance}}, our proposed model has less number of parameters than stacking another linear layer as in 2-MLP$_I$.
% Also compared to 2-MLP$_R$, our model performed better than 2-MLP$_I$ where both model have similar numbers of parameters.
%Not requiring demanding parameters for a task, our method has two advantages against other models when limited samples are available: $1)$ easier to train with less chance of overfitting and 
%$2)$ faster convergence. 
%We dealt the with first advantage 
%We have shown the efficiency of our model in Section \ref{ssec: performance}. 
To show fast convergence of our model, 
we compared the convergence between ours and 2-MLP$_I$ with the same objective function. 
The test accuracy on 4-way (CN vs. EMCI vs. LMCI vs. AD) classification with 5-fold CV is shown in %\ref{fig:convergence_best_model} and 
Fig.~\ref{fig:convergence_same_lr} using 
%$1)$ the best performing models and $2)$ 
% the same learning rate set for both 2-MLP$_I$ and ours. 
the same learning rate set for both models.

% \noindent\textbf{Convergence with the best performing model}
% We brought each best performing setting for baseline (2-MLP$_I$) and Ours($|S|=16$) on 4-way classification with cortical thickness and tau protein. 
% In these two experiments, our methods converged faster than 2-MLP$_I$ in both experiments as shown in \figurename{\ref{fig:convergence_best_model}}.

% \begin{figure}[h!]
% \centering
% \includegraphics[width=\linewidth]{figure/convergence_best_model.png}  
% \caption{\small The test accuracy on 4-way classifications using Cortical Thickness and Tau protein each. The best performing settings for our model and 2-MLP$_I$ are applied respectively. The curve is the median and shaded areas are range of test accuracy over 5-fold runs.}
% \label{fig:convergence_best_model}
% \end{figure}

%\noindent\textbf{Convergence with the Same Learning Rate.}
%In the same task, to be more precise, we set a same learning rate for baseline (2-MLP$_I$) and our proposed model.
As shown in Fig.~\ref{fig:convergence_same_lr}, with same learning rates of 0.1 or 0.01, 
% our proposed model converged faster than baseline model (2-MLP$_I$). 
our proposed model converged faster than 2-MLP$_I$.
To reach the test accuracy of 0.6, our model required 110 and 50 epochs for 0.01 and 0.1 respectively. 
% However, 2-MLP$_I$ required significantly more number of iterations
% and 2-MLP$_I$ with learning rate of 0.1 cannot even reach test accuracy of 0.6 within 1200 epoches showing unstable pattern across different folds with very high variations.
However, 2-MLP$_I$ with learning rate of 0.01 required 430 epochs
and cannot even reach when learning rate was 0.1 showing unstable pattern across 5 folds with very high variations.
% and 2-MLP$_I$ with learning rate of 0.1 cannot even reach test accuracy of 0.6 within 1200 epoches showing unstable pattern across different folds with very high variations.

\begin{figure}[!h]
    \vspace{-5pt}
    \centering
    \includegraphics[width=0.98\linewidth, height = 3.8cm]{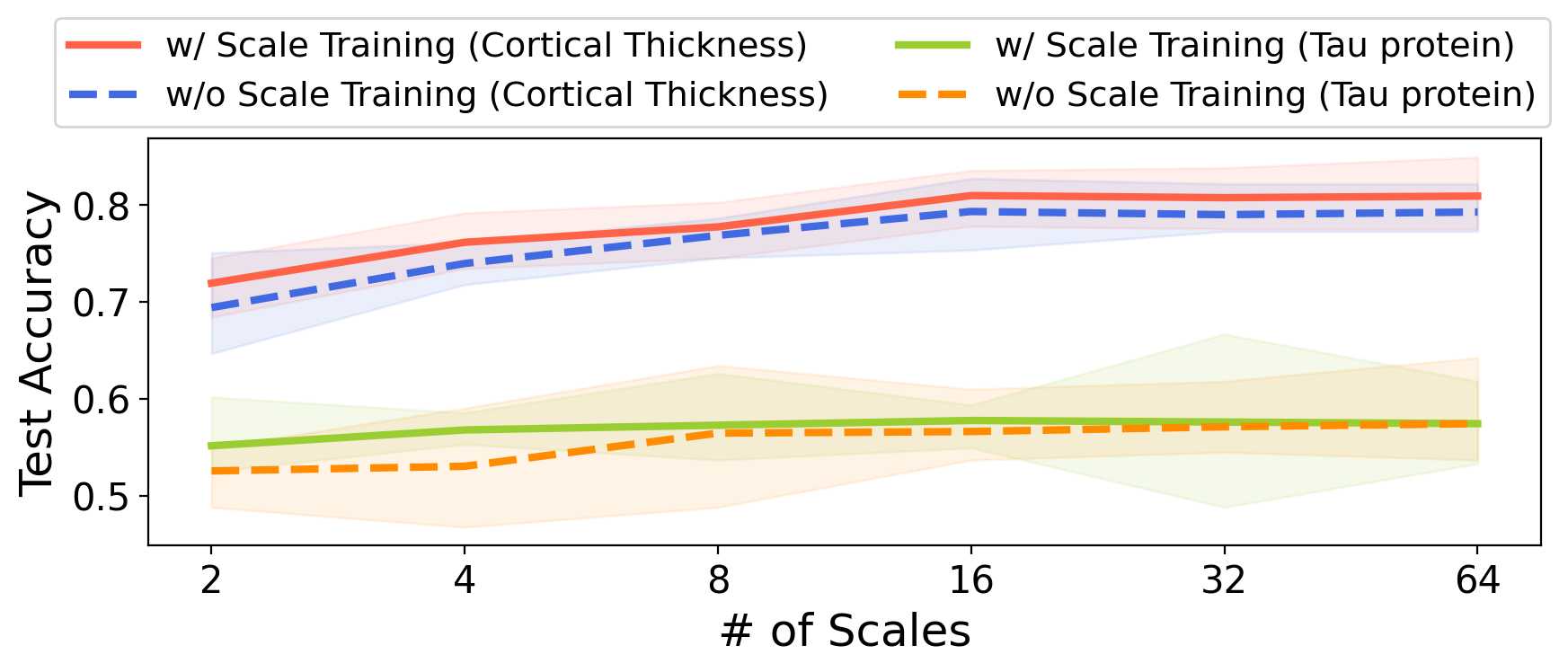}
    \vspace{-10pt} 
    \caption{\small 
    Mean test accuracy w.r.t. scale in our model on 4-way classification using cortical thickness (5-fold CV). 
    Test accuracy improves with scale training (solid line) over without training (dashed line). 
    %The curve is the mean and shaded areas are range of test accuracy over 5-fold runs.
    } 
    \label{fig:effect_kernel_numbers}
    \vspace{-15pt}
\end{figure}

\noindent \textbf{Effect of Number of Scales.}  
% A essential part of our proposed model is trainable kernel defined in a dual space. 
% To figure out the effect of trainable kernels on performances, we divided this into effect of the number of scales and effect of training scales.
% In \tablename{\ref{tab:scale_effect}}, we reported the test accuracy on 4-way classification of our models by $(1)$ varying the number of scales and $(2)$ fixing the scale parameters respectively.
In Section \ref{sec:kernel}, we hypothesized that each  kernel is capable of capturing important frequency components.
To capture various task-specific components with kernels, we need sufficient number of scales.
%\tablename{ \ref{tab:scale_effect}} summarizes 
Fig. \ref{fig:effect_kernel_numbers} shows the test accuracy with respect to the number of scales.  %from our model % model using different number of scales.
%Each test accuracy was measured from the best performing model at each number of scales.
For each experiment, hyperparameters were tuned to obtain the best results. 
%As we use and train more scales until 16, test accuracy kept increasing for both experiments in cortical thickness and tau protein.
%As we use and train more scales until 16, 
The test accuracy kept increasing with respect to the number of scales until $J$$=$$16$ for both cortical thickness and tau. 
After $J$$=$$16$, the performance remained the same or degraded. 
This may be because the data are being mapped to too high-dimensional space with the increase of $J$. 
%this can be interpreted as over-fitting on training samples.

%\subsubsection{Effect of the number of scales}
%\label{sssec:effect_num_scales}

\noindent \textbf{Effect of Training on Scale.}
To show that training on the scales improves the model performance, %is essential part in our method, 
%we compare the results of before-and-after train
we compared the results of training the same model with and without training on $S$. 
%for each number of scales in [2,128], 
%we trained a pair of models with same setting except process of training scales.
% The results with cortical thickness and tau are given in Fig. \ref{fig:effect_kernel_numbers} as well.  
%\tablename{ \ref{tab:scale_effect}}. 
%Training the scales, our model's 
As shown in Fig.~\ref{fig:effect_kernel_numbers}, the classification results on both biomarkers improved with the training of scales. 
However, as more scales were adopted, effect of training on scales decreased.
This may be because the COVLET transform in many scales sufficiently captures 
all the necessary components for classifying AD-specific groups from the beginning. 
%However, as our model adopted more scales, the gap in test accuracy between fixing and training scales was reduced.
%Note that we initialized scale parameters uniformly in $log_{10}$ space, 
%more scale parameters can cover more task-relevant frequency components from the beginning.
%Also, in experiments without training scales, adopting more scales increased the test accuracy.
%However this doesn't mean employing more scales can replace the role of training scales.
%As in \tablename{ \ref{tab:scale_effect}}, there is a side effect of using immense kernels with excessive increase of parameters.
%Based on Fig. \ref{fig:effect_kernel_numbers}, %\tablename{ \ref{tab:scale_effect}}, 
One may argue that simply adopting more scales can replace the training of scales, but such an approach will significantly increase the dimension of a latent space (i.e., curse of dimensionality) and the number of model parameters. 
% %\figurename{ \ref{fig:trained_kernels}} 
% Fig. \ref{fig:trained_kernels} visualizes how each kernel is changed to extract effective frequency component after training. 
% From the initial scales (in dotted lines), the kernel gets shifted the trained ones (in solid lines). 
% Such a behavior lets the model capture better task-specific components in the dual space, 
% which improves the performance of a downstream classifier. 
% %By training the scales, our model can keep the required number of parameters small without the need of employing numerous scales. 

% https://www.ncbi.nlm.nih.gov/pmc/articles/PMC6551438/pdf/gpsych-2018-100005.pdf

% Grad-CAM results here
% TOP-10 ROIs

% ctx\_rh\_S\_temporal\_sup

% sub\_rh\_hippo 

% ctx\_lh\_G\_temp\_sup-G\_T\_transv 

% sub\_lh\_gp ctx\_rh\_S\_temporal\_inf 

% sub\_rh\_thal 

% ctx\_lh\_G\_insular\_short 

% ctx\_lh\_S\_front\_inf 

% ctx\_rh\_G\_temporal\_middle 

% sub\_lh\_amy 

% ctx\_rh\_G\_occipital\_middle 

% ctx\_lh\_S\_cingul-Marginalis 

% ctx\_lh\_S\_oc-temp\_med\_and\_Lingual

% ctx\_rh\_S\_orbital-H\_Shaped

% ctx\_lh\_G\_and\_S\_cingul-Ant

\vspace{-10pt}
\section{CONCLUSION}
\vspace{-3pt}
\label{sec:conclusion}
We proposed an efficiently trainable framework with small sample-sized datasets 
by utilizing trainable parametric kernels and sample covariance structure.
The kernels are defined as band-pass filters in a dual space spanned by eigenvectors of the covariance matrix, 
whose scales are trained with a task-specific loss. 
%With training scales on task loss, 
The training process achieves multi-scale representation that captures task-specific components in the dual space.
Our model is validated on classifications of diagnostic labels of AD (and preclinical AD) 
for performance and convergence, and identifies 
%and figured out crucial 
personalized AD-relevant ROIs supported by other literature.   %\cite{AD_journal, AD_journal2, AD_journal3}

% Stacking excessive layers in DNN results in abominably
% underdetermined system when training samples are limited,
% which is very common in medical applications. In this re-
% gards, we present a framework capable of deriving an ef-
% ficient high-dimensional space with reasonable increase in
% model size. This is done by utilizing a transform (i.e., con-
% volution) that leverage scale-space theory with covariance
% structure. The overall model trains on this transform together
% with a downstream classifier (i.e., Fully Connected layer) to
% capture the optimal multi-scale representation of the origi-
% nal data which correspond to task-specific components in a
% dual space. Experiments on neuroimaging measures from
% Alzheimer’s Disease Neuroimaging Initiative (ADNI) study
% show that our model performs better and converges faster
% than conventional models even when the model size is sig-
% nificantly reduced. The trained model is made interpretable
% using gradient information over the multi-scale transform to
% delineate personalized AD-specific regions in the brain
\section{Acknowledgement}
\vspace{-2mm}
\sloppy This research was supported by HU22C0168, IITP-2022-2020-001461 (ITRC), HU22C0171, NIH R03 AG070701 and NRF-2022R1A2C2092336 and IITP-2019-0-01906 (AI Graduate Program at POSTECH).

\vspace{-2mm}
\bibliographystyle{IEEEbib}
\bibliography{strings,refs}

\end{document}